# Manifold Relevance Determination

**Andreas C. Damianou**                                          ANDREAS.DAMIANOU@SHEFFIELD.AC.UK
Dept. of Computer Science & Sheffield Institute for Translational Neuroscience, University of Sheffield, UK

**Carl Henrik Ek**                                                          CHEK@CSC.KTH.SE
KTH – Royal Institute of Technology, CVAP Lab, Stockholm, Sweden

**Michalis K. Titsias**                                                MTITSIAS@WELL.OX.AC.UK
Wellcome Trust Centre for Human Genetics, Roosevelt Drive, Oxford OX3 7BN, UK

**Neil D. Lawrence**                                             N.LAWRENCE@SHEFFIELD.AC.UK
Dept. of Computer Science & Sheffield Institute for Translational Neuroscience, University of Sheffield, UK

## Abstract

In this paper we present a fully Bayesian latent variable model which exploits conditional non-linear (in)-dependence structures to learn an efficient latent representation. The latent space is factorized to represent shared and private information from multiple views of the data. In contrast to previous approaches, we introduce a relaxation to the discrete segmentation and allow for a "softly" shared latent space. Further, Bayesian techniques allow us to automatically estimate the dimensionality of the latent spaces. The model is capable of capturing structure underlying extremely high dimensional spaces. This is illustrated by modelling unprocessed images with tenths of thousands of pixels. This also allows us to directly generate novel images from the trained model by sampling from the discovered latent spaces. We also demonstrate the model by prediction of human pose in an ambiguous setting. Our Bayesian framework allows us to perform disambiguation in a principled manner by including latent space priors which incorporate the dynamic nature of the data.

## 1. Introduction

Multiview learning is characterised by data which contain observations from several different modalities: for example depth cameras provide colour and depth images from the same scene, or a meeting might be represented by both an audio and a video feed. This motivates latent variable models which align the different views by assuming that a portion of the data variance is shared between the modalities, whilst explaining the remaining variance with latent spaces that are private to each modality. This model structure allows inference when only a subset of the modalities is available and, because the observation spaces have been aligned, it is possible to transfer information between modalities by conditioning the model through the underlying concept.

Several approaches that combine multiple views have been suggested. One line of work aims to find a low-dimensional representation of the observations by seeking a transformation of each view. Different approaches exploit different characteristics of the data such as, correlation (Kuss & Graepel, 2003; Ham et al., 2005), or mutual information (Memisevic et al., 2011). However, these methods only aim to encode the shared variance and do not provide a probabilistic model. To address these shortcomings different generative models have been suggested. In particular, approaches formulated as Gaussian Processes Latent Variable Models (GP-LVMs) (Lawrence, 2005) have been especially successful (Shon et al., 2006; Ek et al., 2007). However, these models assume that a single latent variable is capable of representing each modality, implying that the modalities can be fully aligned. To overcome this, the idea of a factorized latent space was presented in (Ek et al., 2008) where each view is associated with an additional *private* space, representing the variance which cannot be aligned, in addition to the shared space (Ek, 2009), an idea independently suggested by Klami & Kaski (2006). The main challenge for the applicability of the proposed models is that the factorization of the latent variable is a structural and essentially discrete property of the model, making it very challenging to learn. Salzmann et al. (2010) intro-duced a set of regularizers allowing the dimensionality of the factorization to be learned. However, the regularizers were motivated out of necessity rather than principle and introduced several additional parameters to the model.

We present a new principled approach to learning a factorized latent variable representation of multiple observation spaces. We introduce a relaxation of the structural factorization of the model from the original *hard* discrete representation, where each latent variable is either associated with a private space or a shared space, to a smooth continuous representation, where a latent variable may be more important to the shared space than the private space. In contrast to previous approaches the model is fully Bayesian, allowing estimation of both the dimensionality and the structure of the latent representation to be done automatically. Further, it provides an approximation to the full posterior of the latent points given the data. We describe the model and the variational approximation in the next section. The model is capable of handling extremely high dimensional data. We illustrate this by modelling image data directly in the pixel space in section 3. We also demonstrate the model's ability to reconstruct pose from silhouette in a human motion example and, finally, by considering class labels to be a second 'view' of a dataset we show how the model can be used to improve classification performance in a well known visualization benchmark: the "oil data".

## 2. The Model

We wish to relate two views $Y \in \mathbb{R}^{N \times D_Y}$ and $Z \in \mathbb{R}^{N \times D_Z}$ of a dataset within the same model. We assume the existence of a single latent variable $X \in \mathbb{R}^{N \times Q}$ which, through the mappings $\{f_d^Y\}_{d=1}^{D_Y} : X \mapsto Y$ and $\{f_d^Z\}_{d=1}^{D_Z} : X \mapsto Z$ ($Q < D$), gives a low dimensional representation of the data. Our assumption is that the data is generated from a low dimensional manifold and corrupted by additive Gaussian noise $\epsilon^{\{Y,Z\}} \sim \mathcal{N}(\mathbf{0}, \sigma_\epsilon^{\{Y,Z\}} I)$,

$$y_{nd} = f_d^Y(\mathbf{x}_n) + \epsilon_{nd}^Y$$
$$z_{nd} = f_d^Z(\mathbf{x}_n) + \epsilon_{nd}^Z, \quad (1)$$

where $\{y, z\}_{nd}$ represents dimension $d$ of point $n$. This leads to the likelihood under the model, $P(Y, Z|X, \boldsymbol{\theta})$, where $\boldsymbol{\theta} = \{\boldsymbol{\theta}^Y, \boldsymbol{\theta}^Z\}$ collectively denotes the parameters of the mapping functions and the noise variances $\sigma_\epsilon^{\{Y,Z\}}$. Finding the latent representation $X$ and the mappings $f^Y$ and $f^Z$ is an ill-constrained problem. Lawrence (2005) suggested regularizing the problem by placing Gaussian process (GP) (Rasmussen & Williams, 2006) priors over the mappings and the resulting models are known as Gaussian Process latent variable models (GP-LVMs).

In the GP-LVM framework each generative mapping is modeled as a product of independent GP's parametrized by a (typically shared) covariance function $k^{\{Y,Z\}}$ evaluated over the latent variable $X$, so that

$$p(F^Y|X, \boldsymbol{\theta}^Y) = \prod_{d=1}^{D_Y} \mathcal{N}(\mathbf{f}_d^Y|\mathbf{0}, K^Y), \quad (2)$$

where $F^Y = \{\mathbf{f}_d^Y\}_{d=1}^{D_Y}$ with $f_{nd}^Y = f_d^Y(\mathbf{x}_n)$, and similarly for $F^Z$. This allows for general non-linear mappings to be marginalised out analytically leading to a likelihood as a product of Gaussian densities,

$$P(Y, Z|X, \boldsymbol{\theta}) = \prod_{\mathcal{K}=\{Y,Z\}} \int p(\mathcal{K}|F^\mathcal{K}) p(F^\mathcal{K}|X, \boldsymbol{\theta}^\mathcal{K}) \mathrm{d}F^\mathcal{K}. \quad (3)$$

A fully Bayesian treatment requires integration over the latent variable $X$ in equation (3) which is intractable, as $X$ appears non-linearly in the inverse of the covariance matrices $K^Y$ and $K^Z$ of the GP priors for $f^Y$ and $f^Z$. In practice, a maximum a posteriori solution (Shon et al., 2006; Ek et al., 2007; Salzmann et al., 2010) was often used. However, failure to marginalize out the latent variables means that it is not possible to automatically estimate the dimensionality of the latent space or the parameters of any prior distributions used in the latent space. We show how we can obtain an approximate Bayesian training and inference procedure by variationally marginalizing out $X$. We achieve this by building on recent variational approximations for standard GP-LVMs (Titsias & Lawrence, 2010; Damianou et al., 2011). We then introduce *automatic relevance determination* (ARD) priors (Rasmussen & Williams, 2006) so that each view of the data is allowed to estimate a separate vector of ARD parameters. This allows the views to determine which of the emerging private and shared latent spaces are relevant to them. We refer to this idea as *manifold relevance determination* (MRD).

### 2.1. Manifold Relevance Determination

We wish to recover a factorized latent representation such that the variance shared between different observation spaces can be aligned and separated from variance that is specific (private) to the separate views. In manifold relevance determination the notion of a hard separation between private and shared spaces is relaxed to a continuous setting. The model is allowed to (and indeed often does) completely allocate a latent dimension to private or shared spaces, but may also choose to endow a shared latent dimension with more or less relevance for a particular dataview. Importantly, this factorization is learned from data by maximizing a variational lower bound on the model evidence, rather than through construction of bespoke regularizers to achieve the same effect. The model we propose can be seen as a generalisation of the traditional approach to manifold learning; we still assume the existence of a low-dimensional representation encoding the underlying phenomenon, but the variance contained in an observation space does not necessarily need to be governed by the

# Manifold Relevance Determination

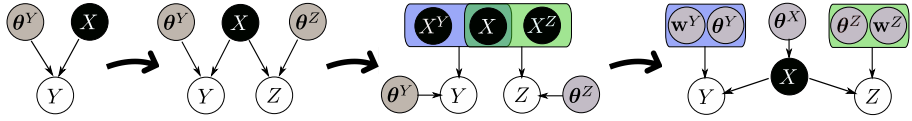

*Figure 1.* Evolution of the structure of GP-LVM model variants. *Far left:* Lawrence (2005)'s original model is shown, a single latent variable $X$ is used to represent the observed data $Y$. Evolved shared models then (*left* to *right*) assume firstly, that all of the variance in the observations was shared (Shon et al., 2006). Secondly, Ek et al. (2008) introduced private latent spaces to explain variance specific to one of the views. MAP estimates used in this model meant the structure of the latent space could not be automatically determined. The rightmost image shows the model we propose in this paper. In this figure we have separated the ARD weights $\mathbf{w}^{\{Y,Z\}}$ from the full set of model hyperparameters $\boldsymbol{\theta}^{\{Y,Z\}} = \{\sigma_\epsilon^{\{Y,Z\}}, \sigma_{ard}^{\{Y,Z\}}, \mathbf{w}^{\{Y,Z\}}\}$, just to emphasize the usage of ARD covariance functions. The latent space $X$ is marginalised out and we learn a distribution of latent points for which additional hyperparamters encode the relevance of each dimension independently for the observation spaces and, thus, automatically define a factorisation of the data. The distribution $p(X) = p(X|\boldsymbol{\theta}^X)$ placed on the latent space also enables the incorporation of prior knowledge about its structure.

full manifold, as traditionally assumed, nor by a subspace geometrically orthogonal to that, as assumed in Salzmann et al. (2010).

The expressive power of our model comes from the ability to consider non-linear mappings within a Bayesian framework. Specifically, our $D_Y$ latent functions $f_d^Y$ are selected to be independent draws of a zero-mean GP with an ARD covariance function of the form:

$$k^Y(\mathbf{x}_i, \mathbf{x}_j) = (\sigma_{ard}^Y)^2 e^{-\frac{1}{2}\sum_{q=1}^Q w_q^Y (x_{i,q}-x_{j,q})^2}, \quad (4)$$

and similarly for $f^Z$. Accordingly, we can learn a common latent space[1] but we allow the two sets of ARD weights $\mathbf{w}^Y = \{w_q^Y\}_{q=1}^Q$ and $\mathbf{w}^Z = \{w_q^Z\}_{q=1}^Q$ to automatically infer the responsibility of each latent dimension for generating points in the $Y$ and $Z$ spaces respectively. We can then automatically recover a segmentation of the latent space $X = (X^Y, X^s, X^Z)$, where $X^s \in \mathbb{R}^{N \times Q_s}$ is the shared subspace, defined by the set of dimensions $q \in [1,...,Q]$ for which $w_q^Y, w_q^Z > \delta$, with $\delta$ being a number close to zero and $Q_s \leq Q$. This equips the model with further flexibility, because it allows for a "softly" shared latent space, if the two sets of weights are both greater than $\delta$ but dissimilar, in general. As for the two private spaces, $X^Y$ and $X^Z$, they are also being inferred automatically along with their dimensionalities $Q_Y$ and $Q_Z$ [2]. More precisely:

$$X^Y = \{\mathbf{x}_q\}_{q=1}^{Q_Y} : \mathbf{x}_q \in X, \ w_q^Y > \delta, \ w_q^Z < \delta \quad (5)$$

and analogously for $X^Z$. Here, $\mathbf{x}_q$ denotes columns of $X$, while we assume that data are stored by rows. All of the above are summarised in the graphical model of figure 1.

## 2.2. Bayesian training

The fully Bayesian training procedure requires maximisation of the logarithm of the joint *marginal* likelihood

$p(Y, Z|\boldsymbol{\theta}) = \int p(Y, Z|X, \boldsymbol{\theta})p(X)\mathrm{d}X$, where a prior distribution is placed on $X$. This prior may be a standard normal distribution or may generally depend on a set of parameters $\boldsymbol{\theta}^X$. By looking again at (3) we see that the above integral is intractable due to the nonlinear way in which $X$ appears in $p(F^{\{Y,Z\}}|X, \boldsymbol{\theta}^{\{Y,Z\}})$. Standard variational approximations are also intractable in this situation. Here, we describe a non-standard method which leads to an analytic solution.

As a starting point, we consider the mean field methodology and seek to maximise a variational lower bound $F_v(q, \boldsymbol{\theta})$ on the logarithm of the true marginal likelihood by relying on a variational distribution which factorises as $q(\Theta)q(X)$, where we assume that $q(X) \sim \mathcal{N}(\boldsymbol{\mu}, S)$. As will be explained later more clearly, in our approach $q(\Theta)$ is a distribution which depends on additional variational parameters $\Theta = \{\Theta^Y, \Theta^Z\}$ so that $q(\Theta) = q(\Theta^Y)q(\Theta^Z)$. These additional parameters $\Theta$ as well as the exact form of $q(\Theta)$ will be defined later on, as they constitute the most crucial ingredient of our non-standard variational approach.

By dropping the model hyperparameters $\boldsymbol{\theta}$ from our expressions, for simplicity, we can use Jensen's inequality and obtain a variational bound $F_v(q) \leq \log p(Y, Z)$:

$$F_v(q) = \int q(\Theta)q(X) \log \left( \frac{p(Y|X)p(Z|X)}{q(\Theta)} \frac{p(X)}{q(X)} \right) \mathrm{d}X$$
$$= \mathcal{L}_Y + \mathcal{L}_Z - \mathrm{KL}\left[q(X) \parallel p(X)\right], \quad (6)$$

where $\mathcal{L}_Y = \int q(\Theta^Y)q(X) \log \frac{p(Y|X)}{q(\Theta^Y)} \mathrm{d}X$ and similarly for $\mathcal{L}_Z$. However, this does not solve the problem of intractability since the challenging terms still appear in $\mathcal{L}_Y$ and $\mathcal{L}_Z$. To circumvent this problem, we follow Titsias & Lawrence (2010) and apply the "data augmentation" principle, i.e. we expand the joint probability space with $M$ extra samples $U^Y$ and $U^Z$ of the latent functions $f^Y$ and $f^Z$ evaluated at a set of pseudo-inputs (known as "inducing points") $\bar{X}^Y$ and $\bar{X}^Z$ respectively. Here, $U^Y \in \mathbb{R}^{M_Y \times D_Y}$, $U^Z \in \mathbb{R}^{M_Z \times D_Z}$, $\bar{X}^Y \in \mathbb{R}^{M_Y \times Q}$, $\bar{X}^Z \in \mathbb{R}^{M_Z \times Q}$ and $M = M_Y + M_Z$. The expression of the joint probability is

---

[1] As we will see in the next section, we actually learn a common *distribution* of latent points.

[2] In general, there will also be dimensions of the initial latent space which are considered unnecessary by both sets of weights.

as before except for the term $p(Y|X)$ which now becomes:

$$p(Y|X, \bar{X}^Y) = \int p(Y|F^Y)p(F^Y|U^Y, X, \bar{X}^Y) \cdot \\ p(U^Y|\bar{X}^Y)\mathrm{d}F^Y \mathrm{d}U^Y \quad (7)$$

and similarly for $p(Z|X)$. The integrations over $U^{\{Y,Z\}}$ are tractable if we assume Gaussian prior distributions for these variables. As we shall see, the inducing points are *variational* rather than model parameters. More details on the variational learning of inducing variables in GPs can be found in Titsias (2009).

Analogously to Titsias & Lawrence (2010), we are now able to define $q(\Theta) = q(\Theta^Y)q(\Theta^Z)$ as

$$q(\Theta) = \prod_{\mathcal{K}=\{Y,Z\}} q(U^\mathcal{K})p(F^\mathcal{K}|U^\mathcal{K}, X, \bar{X}^\mathcal{K}), \quad (8)$$

where $q(U^{\{Y,Z\}})$ are free form distributions. In that way, the $p(F^\mathcal{K}|U^\mathcal{K}, X, \bar{X}^\mathcal{K})$ factors cancel out with the "difficult" terms of $\mathcal{L}_Y$ and $\mathcal{L}_Z$, as can be seen by replacing equations (8) and (7) back to (6), which now becomes our final objective function and can be trivially extended for more than two observed datasets. This function is jointly maximised with respect to the model parameters, involving the latent space weights $\mathbf{w}^Y$ and $\mathbf{w}^Z$, and the variational parameters $\{\boldsymbol{\mu}, S, \bar{X}\}$. As in standard variational inference, this optimisation gives, as a by-product, an approximation of $p(X|Y, Z)$ by $q(X)$, i.e. we obtain a distribution over the latent space. This adds extra robustness to our model, since previous approaches rely on MAP estimates for the latent points. More detailed derivation of the variational bound can be found in the suppl. material.

**Dynamical Modelling:** The model formulation described previously is also covering the case when we wish to additionally model correlations between datapoints of the same output space, e.g. when $Y$ and $Z$ are multivariate timeseries. For the dynamical scenario we follow Damianou et al. (2011); Lawrence & Moore (2007) and choose the prior on the latent space to depend on the observation times $\mathbf{t} \in \mathbb{R}^N$, e.g. a GP with a covariance function $k = k(t, t')$. With this approach, we are also allowed to learn the structure of multiple independent sequences which share some commonality by learning a common latent space for all timeseries while, at the same time, ignoring correlations between datapoints belonging to different sequences.

**Inference:** Given a model which is trained to jointly represent two output spaces $Y$ and $Z$ with a common but factorised input space $X$, we wish to generate a new (or infer a training) set of outputs $Z^* \in \mathbb{R}^{N^* \times D_Z}$ given a set of (potentially partially) observed test points $Y^* \in \mathbb{R}^{N^* \times D_Y}$. This is done in three steps. Firstly, we predict the set of latent points $X^* \in \mathbb{R}^{N^* \times Q}$ which is most likely to have generated $Y^*$. For this, we use an approximation to the posterior $p(X^*|Y^*, Y)$, which has the same form as for the standard Bayesian GP-LVM model (Titsias & Lawrence, 2010) and is given by a variational distribution $q(X, X^*)$. To find $q(X, X^*)$ we optimise a variational lower bound on the marginal likelihood $p(Y, Y^*)$ which has analogous form with the training objective function (6). Specifically, we ignore $Z$ and replace $Y$ with $(Y, Y^*)$ and $X$ with $(X, X^*)$ in (6). In the second step, we find the training latent points $X_{NN}$ which are closest to $X^*$ in the *shared* latent space. In the third step, we find outputs $Z$ from the likelihood $p(Z|X_{NN})$. This procedure returns the set of training points $Z$ which best match the observed test points $Y^*$. If we wish to generate novel outputs, we have to propagate the information recovered when predicting $X^*$. Since the shared latent space encodes the same kind of information for both datasets, we can achieve the above by simply replacing the features of $X_{NN}$ corresponding to the shared latent space, with those of $X^*$.

**Complexity:** As in common sparse methods in Gaussian processes (Titsias, 2009), the typical cubic complexity reduces to $O(NM^2)$, where $N$ and $M$ is the total number of training and inducing points respectively. In our experiments we set $M = 100$. Further, the model scales only linearly with the data dimensionality. Indeed, the Gaussian densities in equation (6) result in an objective function which only involves the data matrices $Y$ and $Z$ in expressions of the form $YY^\top$ and $ZZ^\top$ which are $N \times N$ matrices no matter how many features $D_Y$ and $D_Z$ are used to describe the original data. Also, these quantities are constant and can be precomputed. Consequently, our approach can model datasets with very large numbers of features.

## 3. Experiments

The MRD method is designed to represent multiple views of a data set as a set of factorized latent spaces. In this section we will show experiments which exploit this factorized structure. Source code for recreating these experiments is included as supplementary material.

**Yale faces:** To show the ability of our method to model very high-dimensional spaces our first experiment is applied to the Yale dataset (Georghiades et al., 2001; Lee et al., 2005) which contains images of several human faces under different poses and 64 illumination conditions. We consider a single pose for each subject such that the only variations are the location of the light source and the subject's appearance. Since our model is capable of working with very high-dimensional data, it can be directly applied to the raw pixel values (in this case $192 \times 168 = 32,256$ pixels/image) so that we do not have to rely on image feature extraction to pre-process the data, and we can directly sample novel outputs. From the full Yale database, we con-structed a dataset $Y$ containing the pictures corresponding to all $64$ different illumination conditions for each one of $3$ subjects and similarly for $Z$, for $3$ different subjects. In this way, we formed two datasets, $Y$ and $Z$, each consisting of all $64 \times 3$ images corresponding to a set of three different faces, under all possible illumination conditions, therefore, $Y, Z \in \mathbb{R}^{N \times D}$, $N = 192$, $D = 32,256$. We then aligned the order of the images in each dataset, so that each image $\mathbf{y}_n$ from the first one was randomly set to correspond to one of the $3$ possible $\mathbf{z}_n$'s of the second dataset which are depicted in the same illumination condition as $\mathbf{y}_n$. In that way, we matched datapoints between the two datasets only according to the illumination condition and not the identity of the faces, so that the model is not explicitly forced to learn the correspondence between face characteristics.

The latent space variational means were initialised by concatenating the two datasets and performing PCA. An alternative approach would be to perform PCA on each dataset separately and then concatenate the two low dimensional representations to initialise $X$. We found that both initializations achieved similar results. The optimized relevance weights $\{\mathbf{w}^Y, \mathbf{w}^Z\}$ are visualized as bar graphs in figure 2.

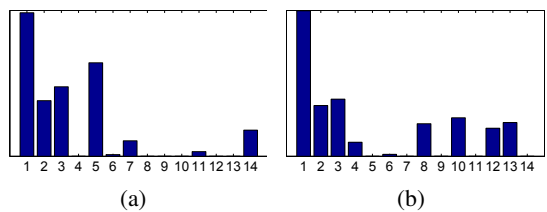

(a)    (b)

*Figure 2.* The relevance weights for the faces data. Despite allowing for soft sharing, the first 3 dimensions are switched on with approximately the same weight for both views of the data. Most of the remaining dimensions are used to explain private variance.

The latent space is clearly segmented into a shared part, consisting of dimensions indexed as $1,2$ and $3$ [3] two private and an irrelevant part (dimension 9). The two data views allocated approximately equal weights to the shared latent dimensions, which are visualized in figures 3(a) and 3(b). Interaction with these three latent dimensions reveals that the structure of the shared subspace resembles a hollow hemisphere. This corresponds to the shape of the space defined by the fixed locations of the light source.

This indicates that the shared space successfully encodes the information about the position of the light source and not the face characteristics. This indication is enhanced by the results found when we performed dimensionality reduction with the standard Bayesian GP-LVM for pictures corresponding to all illumination conditions of a single face

---
[3]Dimension 6 also encodes shared information, but of almost negligible amount ($w_6^Y$ and $w_6^Z$ are almost zero).

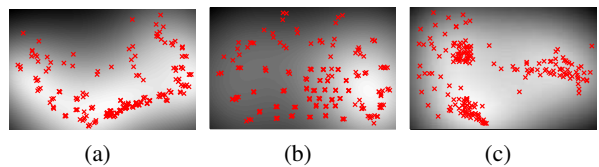

(a)    (b)    (c)

*Figure 3.* Projection of the shared latent space into dimensions $\{1,2\}$ and $\{1,3\}$ (figures (a) and (b)) and projection of the $Y-$private dimensions $\{5, 14\}$ (figure (c)). It is clear how the latent points in figure (c) form three clusters, each responsible for modelling one of the three faces in $Y$.

(i.e. a dataset with one modality). Specifically, the latent space discovered by the Bayesian GP-LVM and the shared subspace discovered by MRD have the same dimensionality and similar structure, as can be seen in figure 4.

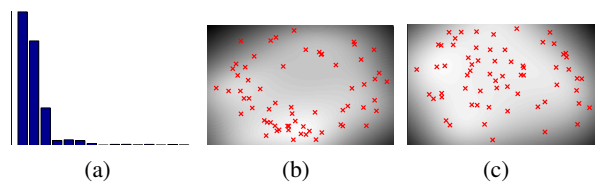

(a)    (b)    (c)

*Figure 4.* Latent space learned by the standard Bayesian GP-LVM for a single face dataset. The weight set $\mathbf{w}$ associated with the learned latent space is shown in (a). In figures (b) and (c) we plotted pairs of the 3 dominant latent dimensions against each other. Dimensions $4, 5$ and $6$ have a very small but not negligible weight and represent other minor differences between pictures of the same face, as the subjects often blink, smile etc.

As for the private manifolds discovered by MRD, these correspond to subspaces for disambiguating between faces of the same dataset. Indeed, plotting the largest two dimensions of the first latent private subspace against each other reveals three clusters, corresponding to the three different faces within the dataset. Similarly to the standard Bayesian GP-LVM applied to a single face, here the private dimensions with very small weight model slight changes across faces of the same subject (blinking etc).

We can also confirm visually the subspaces' properties by sampling a set of novel inputs $X_{samp}$ from each subspace and then mapping back to the observed data space using the likelihoods $p(Y|X_{samp})$ or $p(Z|X_{samp})$, thus obtaining novel outputs (images). To better understand what kind of information is encoded in each of the dimensions of the shared or private spaces, we sampled new latent points by varying only one dimension at a time, while keeping the rest fixed. The first two rows of figure 5 show some of the outputs obtained after sampling across each of the shared dimensions 1 and 3 respectively, which clearly encode the coordinates of the light source, whereas dimension 2 was found to model the overall brightness. The sampling procedure can intuitively be thought as a walk in the space shown in figure 3(b) from left to right and from the bottom to the top. Although the set of learned latent inputs is discrete, the corresponding latent subspace is continuous, and we can interpolate images in new illumination conditions by sampling from areas where there are no training inputs (i.e. in between the red crosses shown in figure 3).

Similarly, we can sample from the private subspaces and obtain novel outputs which interpolate the non-shared characteristics of the involved data. This results in a morphing effect across different faces, which is shown in the last row of figure 5. Example videos can be found in the supplementary material.

As a final test, we confirm the efficient segmentation of the latent space into private and shared parts by automatically recovering all the illumination similarities found in the training set. More specifically, given a datapoint $\mathbf{y}_n$ from the first dataset, we search the whole space of training inputs $X$ to find the 6 Nearest Neigbours to the latent representation $\mathbf{x}_n$ of $\mathbf{y}_n$, based only on the shared dimensions. From these latent points, we can then obtain points in the output space of the second dataset, by using the likelihood $p(Z|X)$. As can be seen in figure 6, the model returns images with matching illumination condition. Moreover, the fact that, typically, the first neighbours of each given point correspond to outputs belonging to different faces, indicates that the shared latent space is "pure", and is not polluted by information that encodes the face appearance.

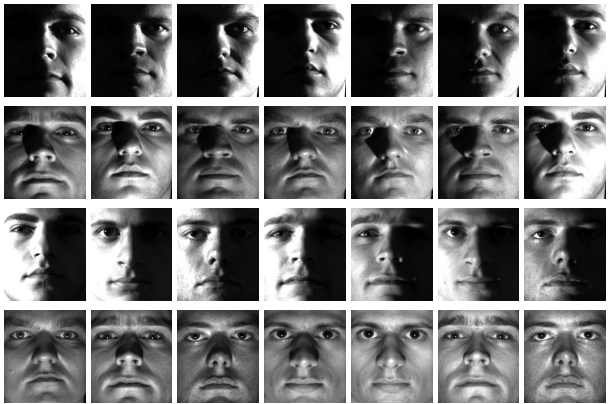

*Figure 6.* Given the images of the first column, the model searches only in the shared latent space to find the pictures of the opposite dataset which have the same illumination condition. The images found, are sorted in columns 2 - 7 by relevance.

**Human motion data:** For our second experiment, we consider a set of 3D human poses and associated silhouettes, coming from the dataset of Agarwal and Triggs (Agarwal & Triggs, 2006). We used a subset of 5 sequences, totalling 649 frames, corresponding to walking motions in various directions and patterns. A separate walking sequence of 158 frames was used as a test set. Each pose is represented by a 63−dimensional vector of joint locations and each silhouette is represented by a 100−dimensional vector of HoG (histogram of oriented gradients) features.

Given the test silhouette features, we used our model to generate the corresponding poses. This is challenging, as the data are multi-modal, i.e. a silhouette representation may be generated from more than one poses (e.g. fig. 7).

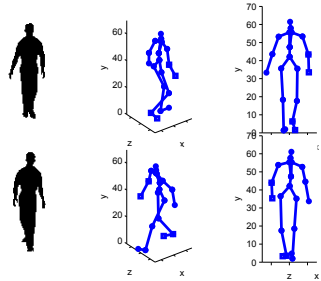

*Figure 7.* Although the two poses in the second column are very dissimilar, they correspond to resembling silhouettes that have similar feature vectors. This happens because the 3D information is lost in the silhouette space, as can also be seen in the third column, depicting the same poses from the silhouettes' viewpoint.

As described in the inference section, given $\mathbf{y}^*$, one of the $N^*$ test silhouettes, our model optimises a test latent point $\mathbf{x}^*$ and finds a series of $K$ candidate initial training inputs $\{\mathbf{x}_{NN}^{(k)}\}_{k=1}^K$, sorted according to their similarity to $\mathbf{x}^*$, taking into account only the shared dimensions. Based on these initial latent points, it then generates a sorted series of $K$ *novel* poses $\{\mathbf{z}^{(k)}\}_{k=1}^K$. For the dynamical version of our model, all test points are considered together and the predicted $N^*$ outputs are forced to form a smooth sequence. Our experiments show that the initial training inputs $\mathbf{x}_{NN}$ typically correspond to silhouettes similar to the given one, something which confirms that the segmentation of the latent space is efficient. However, when ambiguities arise, as the example shown in figure 7, the non-dynamical version of our model has no way of selecting the correct input, since all points of the test sequence are treated independently. But when the dynamical version is employed, the model forces the whole set of training and test inputs to create smooth paths in the latent space. In other words, the dynamics disambiguate the model.

Indeed, as can be seen in figure 8, our method is forced to select a candidate training input $\mathbf{x}_{NN}$ for initialisation which does not necessarily correspond to the training silhouette that is most similar to the test one. What is more, if we assume that the test *pose* $\mathbf{z}^*$ is known and seek for its nearest training neighbour in the pose space, we find that the corresponding silhouette is very similar to the one

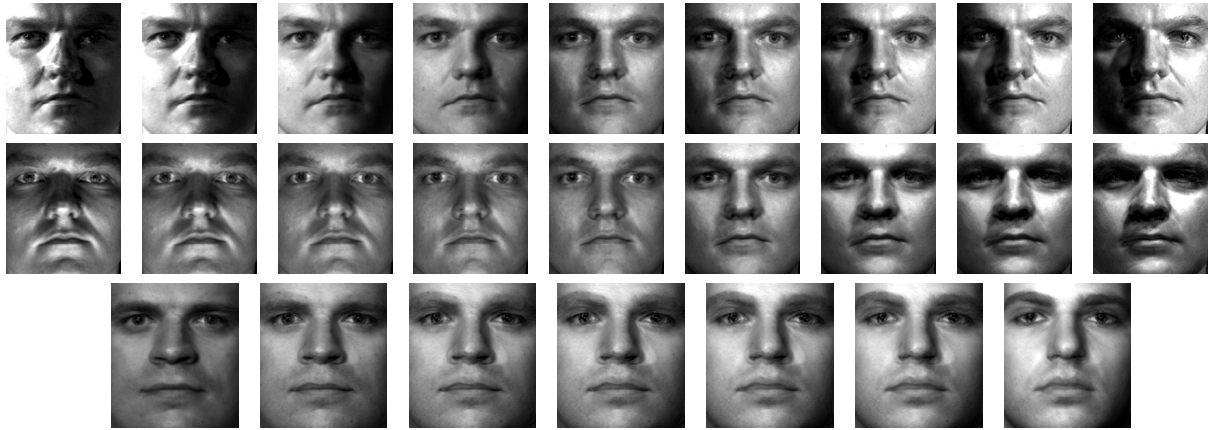

*Figure 5.* Sampling inputs to produce novel outputs. First row shows interpolation between positions of the light source in the $x$ coordinate and second row in the $y$ coordinate (elevation). Last row shows interpolation between face characteristics to produce a morphing effect. Note that these images are presented scaled here, see suppl. material for the original 32,256-dimensional ones.

found by our model, which is only given information in the silhouette space.

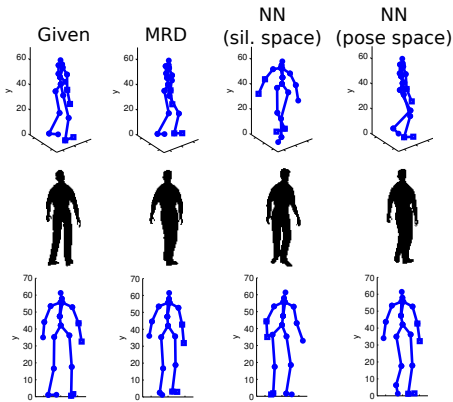

*Figure 8.* Given the HoG features for the test silhouette in column one, we predict the corresponding pose using the dynamical version of MRD and Nearest Neighbour (NN) in the silhouette space obtaining the results in the first row, columns 2 and 3 respectively. The last row is the same as the first one, but the poses are rotated to highlight the ambiguities. Notice that the silhouette shown in the second row for MRD does not correspond exactly to the pose of the first row, as the model generates only a *novel* pose given a test silhouette. Instead, it is the training silhouette found by performing NN in the shared latent space. The NN of the training *pose* given the test pose is shown in column 4.

Given the above, we quantify the results and compare our method with linear and Gaussian process regression and Nearest Neighbour in the silhouette space. We also compared against the shared GP-LVM (Ek et al., 2008; Ek, 2009) which optimises the latent points using MAP and, therefore, requires an initial factorisation of the inputs to be given a priori. Finally, we compared to a dynamical ver-sion of Nearest Neighbour where we kept multiple nearest neighbours and selected the coherent ones over a sequence. The errors shown in table 1 as well as the video provided as supplementary material show that MRD performs better than the other methods in this task.

*Table 1.* The mean of the Euclidean distances of the joint locations between the predicted and the true poses. The Nearest Neighbour in the pose space is not a fair comparison, but is reported here as it provides some insight about the lower bound on the error that can be achieved for this task.

|  | Error |
|---|---|
| Mean Training Pose | 6.16 |
| Linear Regression | 5.86 |
| GP Regression | 4.27 |
| Nearest Neighbour (sil. space) | 4.88 |
| Nearest Neighbour with sequences (sil. space) | 4.04 |
| Nearest Neighbour (pose space) | 2.08 |
| Shared GP-LVM | 5.13 |
| MRD without Dynamics | 4.67 |
| MRD with Dynamics | **2.94** |

**Classification:** As a final experiment, we demonstrate the flexibility of our model in a supervised dimensionality reduction scenario for a classification task. The training dataset was created such that a matrix $Y$ contained the actual observations and a matrix $Z$ the corresponding class labels in 1-of-K encoding. We used the 'oil' database (Bishop & James, 1993) which contains 1000 $12-$dimensional examples split in 3 classes. We selected 10 random subsets of the data with increasing number of training examples and compared to the nearest neighbor (NN) method in the data space. As can be seen in figure 9, MRD successfully determines the shared information between the data and the label space and outperforms NN.

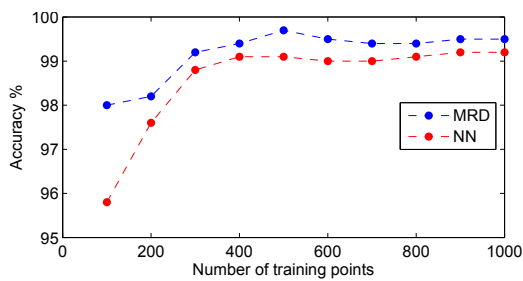

Figure 9. Accuracy obtained after testing MRD and NN on the full test set of the 'oil' dataset.

## 4. Conclusions

We have presented a new factorized latent variable model for multi view data. The model automatically factorizes the data using variables representing variance that exists in each view separately from variance being specific to a particular view. The model learns a distribution over the latent points variationally. This allows us to to automatically find the dimensionality of the latent space as well as to incorporate prior knowledge about its structure. As an example, we showed how dynamical priors can be included on the latent space. This allowed us to use temporal continuity to disambiguate the model's predictions in an ambiguous human pose estimation problem. The model is capable of learning from extremely high-dimensional data. We illustrated this by learning a model directly on the pixel representation of an image. Our model is capable of learning a compact an intuitive representation of such data which we exemplified by generating novel images by sampling from the latent representation in a structured manner. Finally, we showed how a generative model with discriminative capabilities can be obtained by treating the observations and class labels of a dataset as separate modalities.

## Acknowledgments

Research was partially supported by the University of Sheffield Moody endowment fund and the Greek State Scholarships Foundation (IKY). We would like to thank the reviewers for their useful feedback.

## References

Agarwal, Ankur and Triggs, Bill. Recovering 3D human pose from monocular images. 28(1), 2006. doi: 10.1109/TPAMI.2006.21.

Bishop, Christopher M. and James, Gwilym D. Analysis of multi-phase flows using dual-energy gamma densitometry and neural networks. *Nuclear Instruments and Methods in Physics Research*, A327:580–593, 1993.

Damianou, Andreas C., Titsias, Michalis, and Lawrence, Neil D. Variational gaussian process dynamical systems. In *NIPS*, pp. 2510–2518, 2011.

Ek, Carl Henrik. Shared Gaussian Process Latent Variable Models. *PhD Thesis*, 2009.

Ek, Carl Henrik, Torr, Phil, and Lawrence, Neil. Gaussian process latent variable models for human pose estimation. *Proceedings of the 4th international conference on Machine learning for multimodal interaction*, 2007.

Ek, Carl Henrik, Rihan, J, Torr, Phil, Rogez, G, and Lawrence, Neil. Ambiguity modeling in latent spaces. *Machine Learning and Multimodal Interaction*, 2008.

Georghiades, A.S., Belhumeur, P.N., and Kriegman, D.J. From few to many: Illumination cone models for face recognition under variable lighting and pose. *IEEE Trans. Pattern Anal. Mach. Intelligence*, 23(6), 2001.

Ham, J, Lee, D, and Saul, Lawrence K. Semisupervised alignment of manifolds. In *Annual Conference on Uncertainty in Artificial Intelligence*, 2005.

Klami, Arto and Kaski, Samuel. Generative models that discover dependencies between data sets. In *Proceedings of MLSP'06, IEEE International Workshop on Machine Learning for Signal Processing*, pp. 123–128, 2006.

Kuss, Malte and Graepel, Thore. The Geometry Of Kernel Canonical Correlation Analysis. 2003.

Lawrence, Neil. Probabilistic non-linear principal component analysis with Gaussian process latent variable models. *The Journal of Machine Learning Research*, 2005.

Lawrence, Neil D. and Moore, Andrew J. Hierarchical Gaussian process latent variable models. In *ICML*, 2007.

Lee, K.C., Ho, J., and Kriegman, D. Acquiring linear subspaces for face recognition under variable lighting. *IEEE Trans. Pattern Anal. Mach. Intelligence*, 27(5):684–698, 2005.

Memisevic, Roland, Sigal, Leonid, and Fleet, David J. Shared Kernel Information Embedding for Discriminative Inference. *Transactions on Pattern Analysis and Machine Intelligence*, 2011.

Rasmussen, Carl Edward and Williams, Christopher K. I. *Gaussian Processes for Machine Learning*. Cambridge, MA, 2006. ISBN 0-262-18253-X.

Salzmann, Mathieu, Ek, Carl Henrik, Urtasun, Raquel, and Darrell, Trevor. Factorized Orthogonal Latent Spaces. *International Conference on Artificial Intelligence and Statistics*, 2010.

Shon, A, Grochow, K, and Hertzmann, A. Learning shared latent structure for image synthesis and robotic imitation. In *Neural Information Processing*, 2006.

Titsias, Michalis and Lawrence, Neil. Bayesian Gaussian Process Latent Variable Model. In *International Conference on Artificial Intelligence and Statistics*, 2010.

Titsias, Michalis K. Variational learning of inducing variables in sparse Gaussian processes. In *Proceedings of the Twelfth International Workshop on Artificial Intelligence and Statistics*, volume 5, pp. 567–574. 2009.